\documentclass[11pt,a4paper]{article}
\usepackage[hyperref]{emnlp-ijcnlp-2019}
\usepackage{times}
\usepackage{latexsym}

\usepackage{algorithmic}
\usepackage{algorithm}
\usepackage{amsfonts}       
\usepackage{amsmath}
\usepackage{amssymb}
\usepackage{amsthm}
\usepackage{bm}
\usepackage{booktabs}      
\usepackage{color}
\usepackage{enumitem}
\usepackage{float}
\usepackage{graphicx}
\usepackage{hyperref}
\usepackage{microtype}      
\usepackage{multirow}
\usepackage{natbib}
\usepackage{nicefrac}       
\usepackage{sidecap}
\usepackage{subcaption}
\usepackage{tabularx,ragged2e,caption}
\usepackage{times}
\usepackage{url}
\usepackage{wrapfig}

\usepackage{amsmath,amsfonts,bm}

\def\eqref#1{equation~\ref{#1}}

\def\1{\bm{1}}

\DeclareMathAlphabet{\mathsfit}{\encodingdefault}{\sfdefault}{m}{sl}
\SetMathAlphabet{\mathsfit}{bold}{\encodingdefault}{\sfdefault}{bx}{n}

\newcommand{\R}{\mathbb{R}}

\graphicspath{{./images/}}
\DeclareGraphicsExtensions{.pdf,.jpeg,.jpg,.eps,.png}

\usepackage{amsmath, amsthm, amssymb, xspace}

\newcommand{\INST}[1]{}

\usepackage{color}
\definecolor{blue}{rgb}{0,0,.7}
\definecolor{red}{rgb}{.7,0,0}
\definecolor{orange}{rgb}{1,.6,0}
\definecolor{purple}{rgb}{.4,0,.5}
\definecolor{brown}{rgb}{.4,.2,.1}
\definecolor{green}{rgb}{0,.5,0}

\newcommand{\brck}[1]{\left(#1\right)}

\newcommand{\brckcur}[1]{\left\{#1\right\}}

\newcommand{\brcka}[1]{\langle #1\rangle}

\newcommand{\fr}[2]{\frac{#1}{#2}}

\newcommand{\be}{\begin{equation}}
\newcommand{\ee}{\end{equation}}
\newcommand{\bali}{\begin{eqnarray*}}
\newcommand{\eali}{\end{eqnarray*}}
\newcommand{\eq}[1]{\begin{align}#1\end{align}}

\def\enc{\textrm{Enc}}
\def\dec{\textrm{Dec}}
\def\mem{\textrm{Mem}}
\def\lm{P}

\def\CleverName{Sketch-Fill-A-R}

\aclfinalcopy 

\title{Sketch-Fill-A-R: A Persona-Grounded Chit-Chat Generation Framework}

\author{Michael Shum\thanks{Work done as an intern at Salesforce Research.}$\hspace{6pt}{}^1$,
Stephan Zheng${}^2$,  
Wojciech Kry\'sci\'nski${}^2$,\\
\textbf{Caiming Xiong${}^2$, 
Richard Socher${}^2$} \\
  ${}^1$MIT, ${}^2$Salesforce Research \\
  {\small\texttt{mshum@mit.edu}}\\
  {\small\texttt{\{stephan.zheng,kryscinski,cxiong,rsocher\}@salesforce.com}}
  }

\date{}

\begin{document}
\maketitle
\begin{abstract}
Human-like \emph{chit-chat} conversation requires agents to generate responses that are fluent, engaging and consistent.
We propose \CleverName{}, a framework that uses a persona-memory to generate chit-chat responses in three phases. 
First, it generates dynamic \emph{sketch} responses with open slots.
Second, it generates candidate responses by \emph{filling} slots with parts of its stored persona traits.
Lastly, it \emph{ranks} and selects the final response via a language model score.
\CleverName{} outperforms a state-of-the-art baseline both quantitatively (10-point lower perplexity) and qualitatively (preferred by 55\% heads-up in single-turn and 20\% higher in consistency in multi-turn user studies) on the Persona-Chat dataset.
Finally, we extensively analyze \CleverName{}'s responses and human feedback, and show it is more consistent and engaging by using more relevant responses and questions.
\end{abstract}
\section{Introduction}
%
%
\emph{Chit-chat} is a rich domain that challenges machine learning models to express fluent natural language and to successfully interact with other agents. 
Chit-chat stands in contrast to goal-oriented dialogue, such as when a customer has the explicit goal of booking a flight ticket.
When agents communicate, they each have internal state (e.g., their knowledge, intent) and typically have limited knowledge of the state of other agents \cite{chen2017survey}.
As a result, human-like chit-chat requires agents to be fluent, engaging and consistent with what has been said and their persona \cite{zhang2018personalizing}.

These requirements make learning generative chit-chat models a complex task.
First, given an existing conversation history, there may be a large number of valid responses \cite{vinyals2015neural}.
Hence, supervised learning of chit-chat models that cover a large number of topics and styles requires a significant amount of data \cite{zhou2018design}.
Second, as conversations progress and more opportunities for contradiction arise, maintaining consistency becomes more difficult \cite{serban2016generative,serban2017hierarchical}.
Third, engaging chit-chat responses follow conversational structures that are not captured well by perplexity \cite{dinan2019second}. 
Indeed, our human user studies show that both consistency and engagingness are only weakly correlated with perplexity, and fluency is not at all.

\begin{figure}[t]
    \begin{center}
        \includegraphics[width=\linewidth]{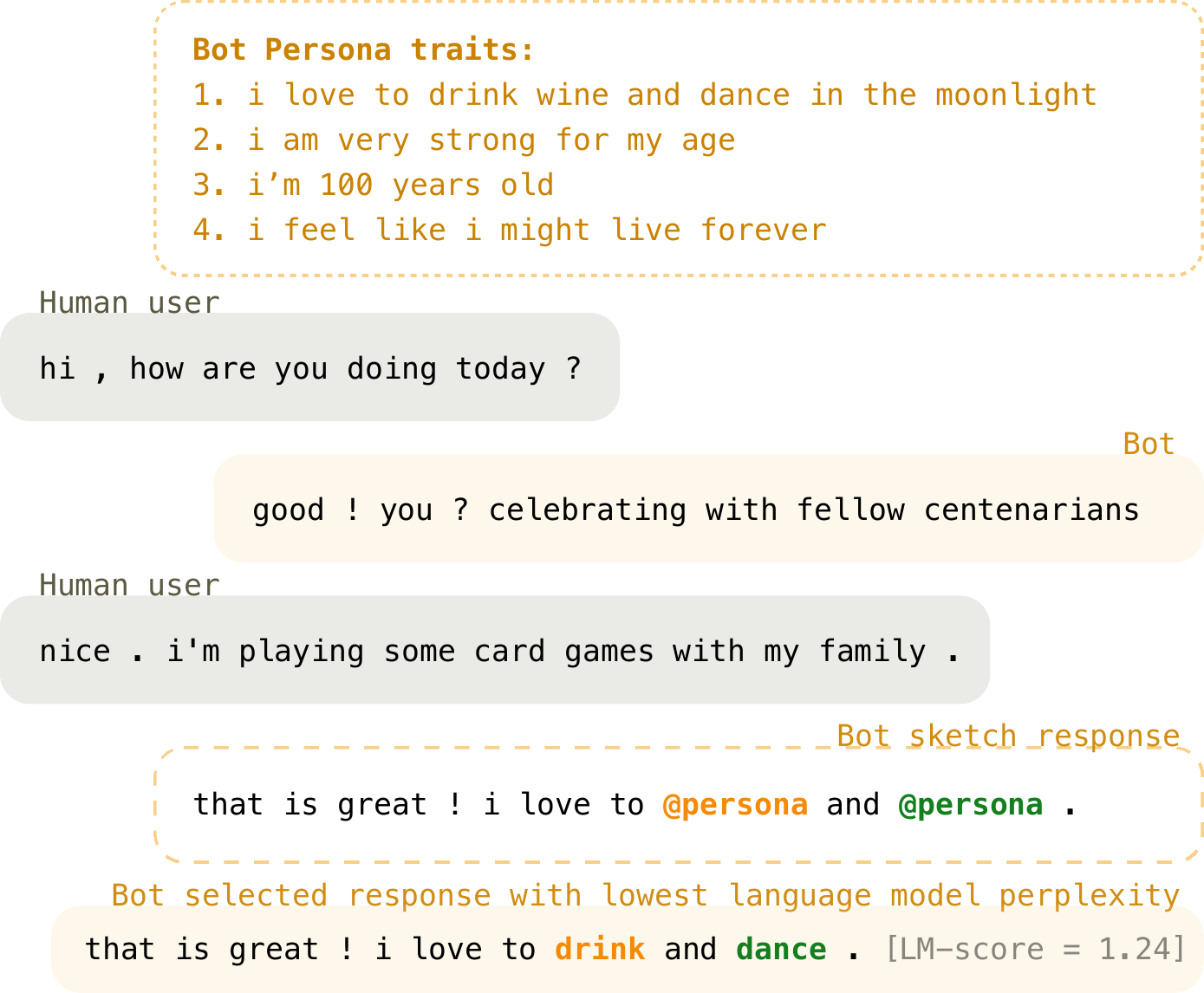}
    \end{center}
    \caption{Chit-chat generation with \CleverName{}.}
    \label{fig:fp}
\end{figure}
%
We propose \CleverName{}, a dialogue agent framework that can learn to generate fluent, consistent and engaging chit-chat responses.
Our key motivation is the hypothesis that human-like chit-chat responses often
1) follow common conversational patterns with insertions of agent-specific traits, and
2) condition explicitly on those persona traits.

\CleverName{} decomposes response generation into three phases: sketching, filling and ranking, see Figure \ref{fig:fp}.
First, \CleverName{} dynamically generates a sketch response with slots, which enables it to learn response patterns that are compatible with many specific persona traits. 
Second, it generates candidate responses by filling in slots with words stored in memory. This enables \CleverName{}'s responses to adhere to its persona. 
Third, the candidate responses are ranked by perplexity under a pre-trained language model (LM), which encourages the final response (with lowest LM perplexity) to be fluent.

%
%
In sum, our contributions are as follows:
\begin{itemize}
    \item We describe \CleverName{} and how its multi-phase generation process encourages fluency, consistency and engagingness.
    \item We show that \CleverName{}  significantly improves hold-out perplexity by $\sim10$ points on the \texttt{Persona-Chat} dataset over state-of-the-art baselines.
    \item We show \CleverName{} is rated higher on conversational metrics and preferred over baselines in single and multi-turn user studies.
    \item We extensively analyze \CleverName{}'s response statistics and human feedback, and show that it is more consistent by using a narrower set of responses, and more engaging, by asking more questions than baselines.
\end{itemize}
\section{Related Work}
\paragraph{Chit-chat Dialogue}
Dialogue agents such as Amazon Alexa, Apple Siri, and Google Home are commonplace today, and are mainly task-oriented: they help users achieve specific tasks. On the other hand, Microsoft XiaoIce \cite{zhou2018design} is an example of an undirected chit-chat dialogue agent. 

Historically task-oriented dialogue systems are composed via components such as dialogue state tracking and natural language generation \cite{jurafsky09}. Even now, the natural language generation component often uses hand-crafted templates and rules defined by domain experts that are filled via heuristics \cite{gao2019neural}. More recently task-oriented dialogue systems have been trained end-to-end \cite{bordes2016learning}, but these systems have specific user intents they aim to fulfill, and so represent a more constrained task.
Early conversational dialogue systems such as ELIZA \cite{weizenbaum1966eliza} and Alice \cite{wallace2009anatomy} were also based on hand-crafted rules and thus brittle. To alleviate this rigidity, more recent neural seq2seq models \cite{sutskever2014sequence} are trained end-to-end \cite{vinyals2015neural,sordoni2015neural,serban2017hierarchical,li2016persona}. 
 To help guide conversation \cite{ghazvininejad2018knowledge,dinan2018wizard} incorporated knowledge-grounded datasets, while \cite{zhang2018personalizing} created the \texttt{Persona-Chat} dataset used in this work. 
\CleverName{} dynamically generates slot sketches and bears resemblance to \cite{wu2018globaltolocal} which assumed data are structured domain-specific triplets and contexts follow templates. 
 However, \CleverName{} does not assume the personas and responses have rigid syntactic structure, and introduces a ranking procedure. 
 Converse to our sketch-and-fill procedure, \cite{qian2017assigning} train a model to select a persona trait and decode around the trait.
 Finally, \cite{welleck2018dialogue} also re-rank by scoring utterances with Natural Language Inference to improve consistency.

\paragraph{Neural Sequence Models}
%
%
\CleverName{} extends a neural encoder-decoder structure \cite{sutskever2014sequence} and is agnostic to the chosen form of encoder-decoder. 
We use recurrent models and attention \cite{bahdanau2014neural}, which auto-regressively embed and generate sequences. 
Hence, our framework is general and is compatible with non-recurrent encoders and decoders, such as Transformer networks with non-recurrent self-attention
\cite{vaswani2017attention,devlin2018bert}.


\CleverName{} uses a simple memory module to store words from personas, which act as context for generation. 
\citet{weston2014memory,sukhbaatar2015end} introduced learned Key-Value Memory Networks, while
\citet{kumar2016ask} introduced Dynamic Memory Nets for question-answering via an iterative attention over memory. 
Also, \CleverName{} decodes responses using a re-ranking strategy based on language model scores, which complements strategies in \cite{kulikov2018importance}.
\begin{figure*}[ht!]
    \begin{center}
    \includegraphics[width=\linewidth]{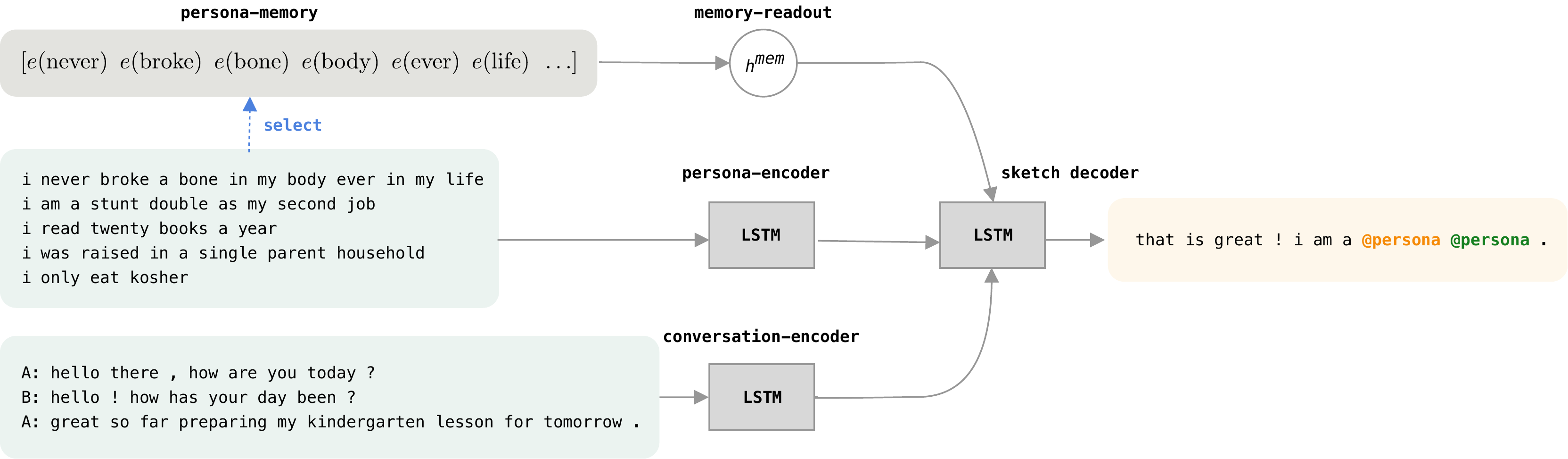}
    \end{center}
    \caption{\CleverName{} generates sketch responses in 4 steps. First, it encodes the conversation history and personas into hidden states $h^{e, \cdot}_t$. It stores word-embeddings for selected rare words from persona traits in a persona-memory. The final encoder hidden state $h_T^{e,c}$ produces a read-out vector $h\textsuperscript{mem}$. Lastly, the decoder outputs a sketch response with \texttt{@persona} slots using $h\textsuperscript{mem}$, encoder hidden states and attention over personas and conversation.}
    \label{fig:ipd100alloc}
\end{figure*}

\begin{figure}[ht!]
    \begin{center}
        \includegraphics[width=0.9\linewidth]{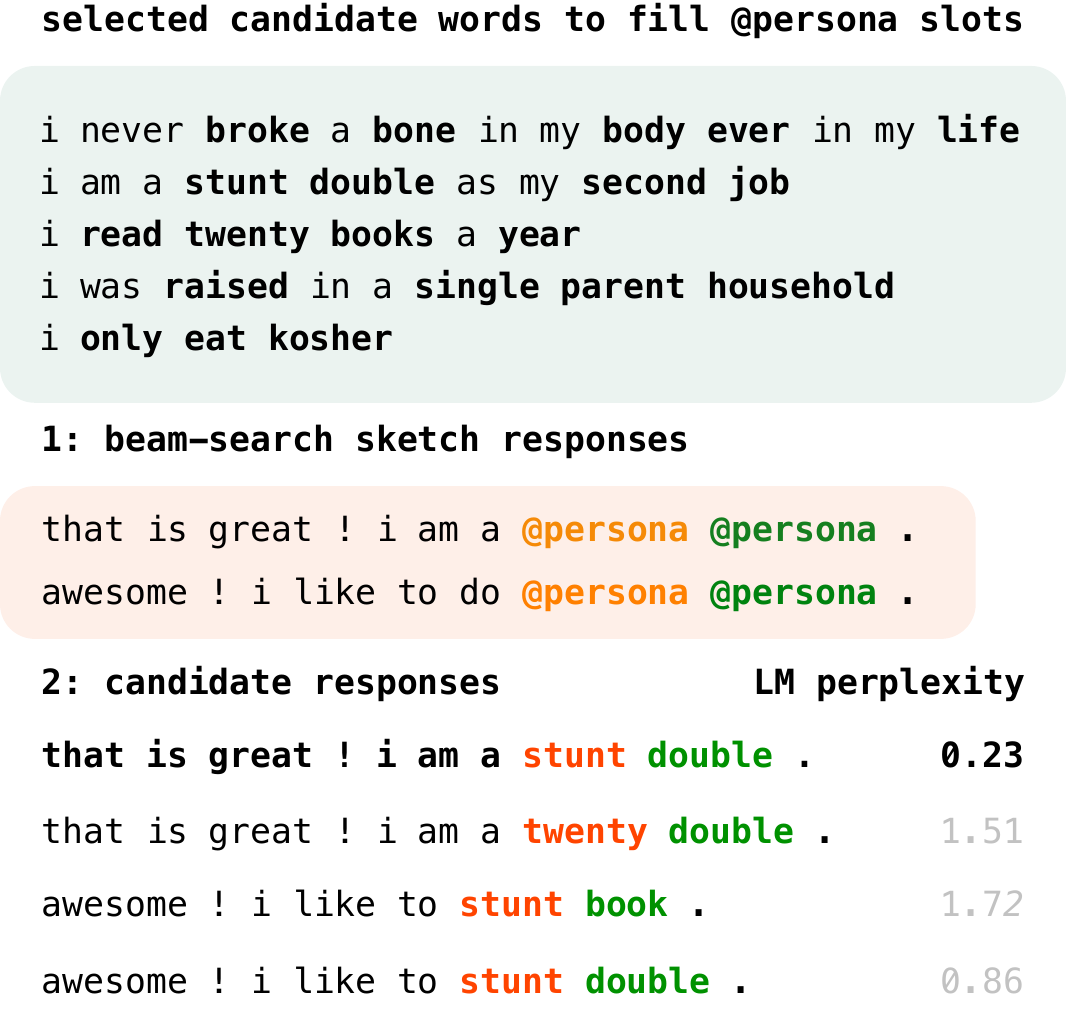}
    \end{center}
    \caption{\CleverName{} inference strategy. During inference, \CleverName{} uses beam search to generate $B$ sketch responses (step 1, depicted $B=2$). In step 2, each beam with \texttt{@persona} slots produces candidate responses by filling it with rare words selected from the persona traits. Finally, a language model scores each candidate and \CleverName{} returns the lowest-perplexity candidate.}
    \label{fig:ipd101alloc}
\end{figure}

\section{\CleverName{}}
Our key motivation is to generate human-like chit-chat responses that are conditioned on persona-relevant information.
\CleverName{} generates chit-chat using a persona-memory to dynamically generate sketches that capture conversational patterns, and inserting persona-relevant information.

To set notation: capitals $W, V, \ldots$ denote matrices, $i,j,k$ are vector-matrix indices and $x, y, \ldots$ denote vectors.
The model input at time $t$ is $x_t$ and the output at time $u$ is $y_u$. 
We denote the conversation by $x^c_t$ and persona trait words by $x^p_t$. 
Both input and output words $x_t, y_u \in \brckcur{0, 1}^V$ are 1-hot vectors, where $V$ denotes the vocabulary size. 
The vocabulary contains all unique words, punctuation and special symbols (e.g., \texttt{EOS}, \texttt{@persona}). $x_{0:T}$ denotes a sequence $(x_0, \ldots, x_T)$.

Formally, we aim to learn a response generation model that predicts words $y_u$ using a probability distribution $P(y_{0:U} | x_{0:T}; \theta)$ over sequences of $T$ words and $N$ persona traits with $R$ rare words. Here $U$ is the output sequence length and $\theta$ are the model weights. We mainly focus on deep neural networks, a model class that has recently seen great success in language generation tasks \cite{sutskever2014sequence,bahdanau2014neural}. 

\CleverName{} composes several components to generate sketch responses: 
\begin{itemize}
    \item An encoder $h^e_{0:T} = \enc\brck{x_{0:T}; \theta}$ that computes hidden representations $e_t$ of the input.
    \item A memory module $h^\textsuperscript{mem} = \mem\brck{x_{0:R}; \theta}$ that stores all rare words from persona traits (constructed by removing stop words). 
    \item A language model $\lm^{LM}\brck{x_{t+1} | x_{0:t}; \theta}$ that computes a distribution over next words. 
    \item A sketch decoder $$h^d_{0:U} = \dec\brck{h^e_{0:T}, h^\textsuperscript{mem}; \theta}$$ that synthesizes both the encoded input and memory readouts, and predicts the next word $\hat{y}_u$ in the sketch response.
\end{itemize}
\subsection{Sketch Response Generation}
\paragraph{Encoder}
We instantiate both encoder and decoder using recurrent neural networks. In this work, we use LSTMs \cite{hochreiter1997long}, although other choices are possible \cite{elman1990finding,2014arXiv1406.1078C}. The encoder computes hidden states $h_{0:T} \in \R^{d\textsubscript{hid}}$ auto-regressively:
\eq{
    h^e_{t+1} &= \textrm{LSTM}\brck{e(x_{t}), h^e_{t}; \theta},
}
where $e(x_t)$ are word-embedding representations of the raw input tokens $x_t$. 
As such, \CleverName{} encodes both conversation history $x^c_{0:T}$ and individual persona traits $x^{p}_{0:T}$ into hidden states $h^\textsuperscript{conv}_{0:T}$ and $h^\textsuperscript{pers}_{0:T}$. We denote final hidden states for all $N$ personas as $h^\textsuperscript{pers}_{0:N}$.
\paragraph{Memory Module}
\CleverName{} selects a subset of rare words, $x^{p}_{r}$ from the persona traits by removing stop-words, punctuation, and other symbols. After encoding the input dialogue, \CleverName{} does a memory readout using the final conversation encoder hidden state $h^\textsuperscript{conv}_{T}$ as a query:
\eq{
    h^\textsuperscript{mem} &= h^\textsuperscript{conv}_T + o, \\
    o &= \sum_r p_{r} x^{p}_{r}C^2, \\
    p_r &= \sigma(((h_{r}^c)^T x^{p}_{r}C^1))
}
where $r$ is a vector index over the rare word memory, $\sigma$ is a softmax activation function creating attention weights $p_i \in \R^{d\textsubscript{hid}}$, and $C^k$ are trainable embedding matrices where $C^k \in \R^{V \times d\textsubscript{hid}}$.
\paragraph{Attention Decoder}
The decoder is an LSTM which recursively computes hidden states $h^d_u$ that are mapped into a distribution over output words:
\eq{
    & h^d_{u+1} = \textrm{LSTM}\brck{ y_{u}, h^d_{u}; \theta}, \label{eq:decrnn}\\    
    & h^d_0 = f\brck{ W^{d} [h^e_T , h^\textsuperscript{mem}] + b^{d}}, \\
    & \lm\brck{y_{u+1} | x_{0:T}, y_{0:u}} = \sigma( c_u W^{\textrm{emb}} ). \label{eq:sketchprob}
}
At decoding time $u+1$ the decoder computes the next hidden state $h^d_{u+1}$ using the previous predicted word $y_u$ and decoder hidden state $h^d_u$, in addition to attention over the context of the response, i.e., previous utterances and the agent's persona traits. 
$W^{d} \in \R^{2*d\textsubscript{hid} \times d\textsubscript{hid}}$ projects $[h^e_T , h^\textsuperscript{mem}]$ down to the initial hidden state of the decoder and $W\textsuperscript{emb} \in \R^{d\textsubscript{hid} \times V}$ is the transpose of the encoding embedding matrix. The decoding context $c_u$ augments decoder hidden state $h^d_u$ with attention vectors $c^{\text{conv}}_{u}$ over encoded hidden states $h\textsuperscript{conv}$ and $c^{\text{pers}}_{u}$ over encoded persona hidden states $h^\text{pers}_{0:N}$:
\eq{
    & c_u   = f\brck{W^{ac}[h^d_u, c^{\text{conv}}_{u}, c^{\text{pers}}_{u}] + b^{ac}}, \\
    & c^{\text{conv}}_{u}   = \brcka{ \sigma( \brcka{ W^a [h^d_u] + b^a, h^\text{conv}_\text{0:T} } ), h^\text{conv}_\text{0:T}}, \label{eq:attnweight} \\
    & c^{\text{pers}}_{u}   = \brcka{ w_{u,n}, h^\text{pers}_\text{0:N}},  \\
    & w_{u,n} = \sigma(\brcka{ W^a [h^d_u] + b^a, h^\text{pers}_\text{0:N} }) \label{eq:persweight}
}
where $f$ is a tanh, $W^{ac} \in \R^{3*d\textsubscript{hid} \times d\textsubscript{hid}}$, $W^a \in \R^{d\textsubscript{hid} \times d\textsubscript{hid}}$ and $\sigma$ is the softmax activation function.
In Equations \ref{eq:attnweight} and \ref{eq:persweight} the softmax is over the encoder time dimension and $\brcka{\cdot, \cdot}$ is an inner product.
\subsection{Inference Reranking Strategy}
\label{sect:inference}
\CleverName{} trains the sketch-decoder outputs (Equation \ref{eq:sketchprob}) by minimizing cross-entropy loss with ground truths $y^*_u$.
However, during inference, \CleverName{} uses an iterative generate-and-score approach to produce the final response:
%
%
\begin{enumerate}
    \item Perform beam search with beam size $B$ to generate $B$ sketch responses $\brckcur {\hat{y}^b_{0:Ux^b} }_{b=1,\ldots, B}  $ that may contain \texttt{@persona} tags.
    \item For each sketch with tags, select the persona $i^*$ with the highest attention weight $w_{u^*,i^*}(h^c_T)$ from the first sketch tag location $u^*$, and construct $B'$ candidate responses by filling each \texttt{@persona} slot with words selected from $i^*$.
    \item Compute the perplexity $s_b$ of all $B'$ candidate responses using a pre-trained language model:
    \eq{
        s_k = \exp\fr{1}{T^b}\sum_{t=0}^{T^b} -\log \lm^{LM}\brck{y^b_u | y^b_{0:u-1}}. \nonumber
    }
    \item The final response is the response $b^* = \min_b s_b$ with the lowest LM-likelihood score.
\end{enumerate}

For models that do not use reranking to fill slots, we follow the methodology of \cite{wu2018globaltolocal} in using a global-to-local memory pointer network in order to fill slots. For detail, see the Appendix. 
\begin{table}[t!]
    \small
    \begin{center}
        \begin{tabular}{ lcc } 
        \small
        \textbf{Model} & \textbf{Parameters} & \textbf{Perplexity} \\ 
        KVMemNet & 46.3M & $34.54$ \\ 
        Sketch-Fill & 26.6M & $26.75$ \\
        Sketch-Fill-R & 26.6M & $26.74$ \\  
        Sketch-Fill-A & 26.9M & $24.17$ \\
        \CleverName{} & 26.9M & $\bm{24.99}$ \\ 
        \end{tabular}
    \caption{\CleverName{} achieves significantly lower out-of-sample perplexity than KVMemNet. Perplexity for \CleverName{} is measured over the sketch template responses. The number of persona tag occurrences is very small, constituting 8\% of the total words. See Appendix for more information.} 
    \label{tab:hyperp}
    \resizebox{\linewidth}{!}{%
        \begin{tabular}{ lcc } 
        \small
        \textbf{Sequence size} & \textbf{KVMemNet} & \textbf{\CleverName{} (ours)} \\     
        Unigram & 5.39\%  & 1.72\% \\
        Bigram & 32.65 \% & 7.32 \% \\
        Trigram & 54.95 \% & 13.97 \% \\
        Full responses & 70.16 \% & 50.60 \% \\
        \end{tabular}
    }
    \caption{Percentage of novel $n$-grams and full responses generated by the KVMemNet and \CleverName{} models computed on the full validation set.}
    \label{tab:ngram-counts}
    \end{center}
\end{table}

\section{Empirical Validation}
To validate \CleverName{}, we first show that it achieves better supervised learning performance than baselines on a chit-chat dialogue dataset. 
\paragraph{\texttt{Persona-Chat} Dataset} 
We trained \CleverName{} to generate single-turn agent responses on the \texttt{Persona-Chat} dataset \cite{zhang2018personalizing}, which contains 10,907 dialogues. 
Here, a dialogue consists of multiple \emph{turns}: a single turn contains the utterance of a single agent.
We processed this dataset into training examples that each consist of the conversation history $x^c_t$, set of persona traits $x^p_t$ of the model, and the ground truth sketch response $y_u$. 
This process yielded 131,438 training examples. 
Rare words were identified by removing all punctuation and stop words from the set of persona traits (see Appendix for more information). Ground truth sketch responses were then constructed by replacing all rare word instances in ground truth responses with \texttt{@persona} tags. 
\paragraph{Language Model Pre-training}
\CleverName{} uses a Transformer-based GPT-network \cite{radford2018improving} pre-trained on the Books text corpus \cite{moviebook} to rank candidate responses with filled \texttt{@persona} slots according to their LM-perplexity scores.
For model details, see the Appendix.
\paragraph{Experimental Setup}
We compared 4 variations of \CleverName{} with a strong baseline: \footnote{
A number of chit-chat models posted results in the ConvAI2 competition. However, we could not reproduce these, as all competitive methods rely on extensive pre-training with large models, or do not have code or trained models available. 
}
\begin{itemize}
    \item Key-Value Memory Network (KVMemNet) \cite{zhang2018personalizing},
    \item Sketch-Fill (SF)
    \item Sketch-Fill-A: SF + attention
    \item Sketch-Fill-R: SF + reranking
    \item \CleverName{}: SF + attention + reranking
\end{itemize}
\cite{zhang2018personalizing} showed not only that models trained on Persona-Chat outperform models trained on other dialogue datasets (movies, Twitter) in engagingness but also that KVMemNet outperforms vanilla Seq2Seq on Persona-Chat. As a result we omit comparison with vanilla Seq2Seq. Further KVMemNet is the strongest of the few public baselines available to compare against on chitchat with personas. 

All \CleverName{} models use language model reranking (see Section \ref{sect:inference}). 
All input tokens $x^c_t, x^p_t$ were first encoded using 300-dimensional GLoVe word embeddings $e(x_t)$ \cite{pennington2014glove}. 
All models were trained by minimizing loss on the ground truth sketch response $y^*_{0:U}$:
\eq{
    \min_{\theta} -\sum_{u=0}^{U} \brcka{y_u^*, \log\lm\brck{y_{u} | x_{0:T}, y_{0:u-1}; \theta} }.
}
For training details, see the Appendix.
The results are shown in Table \ref{tab:hyperp}. Sketch-Fill models outperform KVMemNet on validation perplexity, while using significantly fewer weights than KVMemNet. 
This suggests the structure of Sketch-Fill models fits well with chit-chat dialogue. 
\begin{table*}[t!]
    \begin{center}
    \small
    \resizebox{\linewidth}{!}{%
        \begin{tabular}{ llccc|lccc } 
         & \textbf{Baseline} & \textbf{Consistency} & \textbf{Engagingness} & \textbf{Fluency} & \textbf{Ours} & \textbf{Consistency} & \textbf{Engagingness} & \textbf{Fluency} \\ 
        Test I   & KVMemNet   & $\bm{3.60 \pm 0.84}$ & $\bm{3.81 \pm 0.66}$ & $\bm{4.49 \pm 0.45}$ & Sketch-Fill & $2.51 \pm 1.16$ & $2.57 \pm 1.10$ & $2.98 \pm 1.29$ \\
        Test II  & KVMemNet   & $\bm{3.57 \pm 0.86}$ & $\bm{3.77 \pm 0.62}$ & $\bm{4.54 \pm 0.47}$ & Sketch-Fill-A & $2.49 \pm 1.04$ & $2.51 \pm 1.03$ & $2.75 \pm 1.20$ \\
        Test III & KVMemNet   & $3.18 \pm 1.16$ & $3.51 \pm 0.85$ & $4.41 \pm 0.48$  & Sketch-Fill-R & $\bm{3.34 \pm 1.02}$ & $\bm{3.89 \pm 0.79}$ & $\bm{4.45 \pm 0.78}$ \\
        Test IV  & KVMemNet   & $3.31 \pm 1.03$ & $3.56 \pm 0.78$ & $4.43 \pm 0.48$  & \CleverName{} & $\bm{3.54 \pm 1.01}$ & $\bm{3.69 \pm 0.92}$ & $4.43 \pm 0.71$ \\
        \end{tabular}
        }    
    \end{center}
    \caption{
    User study ratings of single-turn responses (score range 1 (lowest) - 5 (highest)). 
    Each experiment showed generated responses from a \CleverName{}-variation and KVMemNet on 100 conversations to 5 human raters.
    %
    %
    Each row shows ratings from a single heads-up experiment. 
    Sketch-Fill with reranking show a small gain over KVMemNet on all qualitative metrics, but the variance in the ratings is high. 
    Sketch-Fill without reranking perform much worse, due to their responses not being fluent, despite achieving low perplexity (see Figure \ref{tab:hyperp}).
    }
    \label{tab:human-study-traits}
\end{table*}
\begin{table}[t!]
    \begin{center}
    \small
    \resizebox{\linewidth}{!}{%
    \begin{tabular}{ lcc } 
        \small
        \textbf{A/B Experiment} & \textbf{KVMemNet} & \textbf{Sketch-Fill-$x$ (ours)}  \\ 
        vs Sketch-Fill   & \textbf{380} & 120 \\
        vs Sketch-Fill-A & \textbf{396} & 103 \\
        vs Sketch-Fill-R & 225 & \textbf{275} \\
        vs \CleverName{} & 232 & \textbf{266} \\
        \end{tabular}
        }    
    \caption{Human user A/B-preferences on 100 conversations, each shown to 5 users. 
    Two \CleverName{} variations are preferred over KVMemNet.
    } 
    \label{tab:human-study-preference}
\end{center}
\end{table}
\begin{table}[t!]
    \begin{center}
    \small
    \resizebox{\linewidth}{!}{%
        \begin{tabular}{ lcccc } 
        \textbf{} & \textbf{Fluency} & \textbf{Consistency} & \textbf{Engagingness} & \textbf{Perplexity} \\ 
        \textbf{Fluency}      & 1 &  0.40 & 0.46 & -0.01 \\
        \textbf{Consistency}  & -   &  1   & 0.67 & -0.20 \\
        \textbf{Engagingness} & -   &  -     & 1   & -0.15 \\
        \textbf{Perplexity}   & -   &  -     & -     & 1     \\
        \end{tabular}
        }    
    \caption{Pearson's correlation $\rho$ between human ratings and perplexity of user study examples. For visual KDE-plots of the data, see the Appendix.} 
    \label{tab:correlations}
\end{center}
\end{table}

\section{User Study and Qualitative Analysis}
%
%
%
Although Sketch-Fill models perform well quantitatively, a crucial test is to evaluate how well they perform when judged by human users on conversational \emph{quality}, which is not explicitly captured by perplexity. 
We performed single and multi-turn dialogue user studies to assess the quality of \CleverName{}, rated along several dimensions: 
\begin{itemize}
    \item \textbf{Fluency}: whether responses are grammatically correct and sound natural.
    \item \textbf{Consistency}: whether responses do not contradict the previous conversation.
    \item \textbf{Engagingness}: how well responses fit the previous conversation and how likely the conversation would continue.
\end{itemize}
Our definition of engagingness includes \emph{relevance}, defined in pragmatics and relevance theory \citep{wilson2002relevance, grice1991studies} as a statement leading to positive cognitive effect. However an engaging statement may be ironic \citep{sperber1981irony}, humorous, or further specific to individuals.

We also explore which qualities of \CleverName{}'s outputs are correlated with human ratings and perplexity scores.
%
%
Our results suggest that:
\begin{itemize}
\item Conditioning on persona-memory provides more consistency.
\item \CleverName{} poses more questions, which correlates with higher engagingness.
\item Responses need to be fluent in order to be consistent or engaging. In addition, more consistent responses are more likely to be engaging.
\item Perplexity is not correlated with high-quality responses. 
\end{itemize}

\begin{table*}[ht!]
\centering
\begin{minipage}{0.48\linewidth}
\small
\begin{tabular}{l}
Model Trait 1: I married a super model from Italy \\
Model Trait 2: I've zero family that I'm close to \\
Model Trait 3: My name is George                      \\
Model Trait 4: I'm a bee farmer                   \\
Model Trait 5: My favorite food is papaya         \\
\hline 
\textbf{Human:}  hi what’s up\\
\textbf{Model:}  do you like papaya ?\\
\textbf{Human:}  eh, i`m more into mango\\
\textbf{Model:}  what is your favorite papaya ?\\
\textbf{Human:}  probably just the normal ones.\\
\textbf{Human:}  what do you like to do?\\
\textbf{Model:}  i am a bee farmer .\\
\textbf{Human:}  woah , that`s cool\\
\textbf{Model:}  how are you ?\\
\textbf{Human:}  i`m doing well , \\
\textbf{Human:}  just trying to get some work done\\
\textbf{Model:}  do you have any hobbies ?\\
\textbf{Human:}  i like to play basketball , what about you?\\
\textbf{Model:}  i love papaya food .\\
\textbf{Human:}  what is your name?\\
\textbf{Model:}  george . what is your favorite name ?\\    
\end{tabular}
\caption[Table caption text]{ Example multi-turn dialog between a human and Sketch-Fill-AR. Model persona traits are shown above; the human was not given specific persona traits.}
\label{tab:multiturn-sketchfillar}
\end{minipage}
\hspace{0.01\linewidth}
\begin{minipage}{0.48\linewidth}
\small
\begin{tabular}{l}
Model Trait 1: I work for our local supermarket \\
Model Trait 2: My favorite band is The Who\\
Model Trait 3: I have never been out of Ohio\\
Model Trait 4: My favorite food is pizza with black olives\\
\hline
\textbf{Human:}  what are you doing today?   \\
\textbf{Model:}  i am ok what do you do for fun ?\\
\textbf{Human:}  I like biking. Do you bike?.\\
\textbf{Model:}  cool ! i am watching a movie with my little girl\\
\textbf{Human:}  You have a daughter? How old is she?.   \\
\textbf{Model:}  oh that is cool . i do not have any myself . \\
\textbf{Human:}  Oh. do you have a girlfriend then?.      \\
\textbf{Model:}  she is an ex singer from the early 2000s ,  \\
\textbf{Model:}  mostly does pop songs and the like       \\
\textbf{Human:}  What kind of movies do you like?       \\
\textbf{Model:}  little girl and my mother lives with us  \\
\textbf{Model:}  she`s a concert pianist, she in between jobs   \\
\textbf{Human:}  who is this little girl?\\
\textbf{Model:}  she also likes karate\\
\textbf{Model:}  she is why i got a brown belt ! \\
\textbf{Human:}  what is her name?\\
\textbf{Model:}  she had a green thumb and loved flowers
\end{tabular}
\caption[Table caption text]{ Example multi-turn dialog between a human and KVMemNet. Model persona traits are shown above; the human was not given specific persona traits.}
\label{tab:multiturn-kvmemnet}
\end{minipage}
\end{table*}

\begin{table}[ht!]
    \begin{center}
    \small
    \resizebox{\linewidth}{!}{%
        \begin{tabular}{ lcc } 
         \textbf{Metric} & \textbf{KVMemNet} & \textbf{\CleverName{} (ours)} \\
            Fluency     & $\bm{3.27 \pm 1.19}$ & $2.83 \pm 1.31$  \\
            Consistency & $2.15 \pm 1.18$ & $\bm{3.72 \pm 1.26}$  \\
            Engagement  & $1.93 \pm 0.92$ & $\bm{2.41 \pm 1.33}$ 
        \end{tabular}
        }    
    \end{center}
    \caption{
    Multi-turn user study ratings (score range 1 (lowest) - 5 (highest)). 
    We collected $30$ conversations with 20 turns between human raters and models.
    KVMemNet is more fluent, but \CleverName{} is more engaging and significantly more consistent.
    }
    \label{tab:multiturn-human-study}
\end{table}

\begin{table}
\small
\begin{tabular}{l}
Human: hi there . how are you \\
Model: hi good just writing some music and you\\
Human: i am well . just took my dog for a walk\\
\hline
\textbf{KVMemNet:}  sorry , i trying to become the next big thing \\ in music , i sing\\
\textbf{\CleverName{}:}  what kind of music do you like to do ?\\
\end{tabular}
\caption{An example single-turn response for KVMemNet and \CleverName{} given the same context.}
\label{tab:singleturn-human-study}
\end{table}

\subsection{Single-turn Experiments}
The studies were completed on 100 random examples sampled from the validation set, where each example was rated by 5 judges.
Judges hired for the study came from English speaking countries. 
As a calibration step, they were shown examples of good and bad responses in all of the measured dimensions, before proceeding with the study.

The study was executed in two settings, fine-grained, where the judges were asked to rate the responses on a scale from 1 (lowest) to 5 (highest) for each of the mentioned dimensions, and binary, where they were asked to choose a response that would best fit the conversation.

The results of the fine-grained survey are presented in Table \ref{tab:human-study-traits}, where each row corresponds to a separate heads-up experiments in which the KVMemNet model was paired with one of the versions of \CleverName{}. 
The study showed small gains on all metrics for all \CleverName{} variations, however, the variance of results was high. 
We believe that this artifact could be caused by a number of factors, including subjective preferences of raters and potential ambiguities in the experiments description. We notice that Sketch-Fill and Sketch-Fill-A reach lower perplexity values than KVMemNet, but comparatively have lower evaluations across the board. Conversely, ranking models like Sketch-Fill-R and Sketch-Fill-A-R have higher scores on all metrics. We observe that the difference is due to the ranker giving more fluent outputs via better selection of persona words to use.

Table \ref{tab:human-study-preference} shows the results of the human study in a binary setting. 
In these experiments the base and attention-augmented versions of \CleverName{} outperformed KVMemNet by a clear margin.

The following subsections present in-depth analysis of the human study. The analysis focuses on the \CleverName{} model, since it yielded the best perplexity and user study results.


\paragraph{Correlation between ratings} 
To study and better understand the reasoning behind the ratings assigned by annotators, we look at the correlation between the different dimensions in which responses where scored.
Figure \ref{tab:correlations} shows Kernel-Density-Estimation plots of the data points and associated Pearson correlation coefficients $\rho$.
The data shows weak ($\rho = 0.397$) to moderate ($\rho = 0.462$) correlation between \textit{fluency} and \textit{consistency}, and \textit{fluency} and \textit{engagingness} ratings respectively. 
The data shows $\rho$ value of $0.670$ between \textit{engagingness} and \textit{consistency} ratings, suggesting strong correlation between those dimensions. See appendix for more detailed information.
The numbers were obtained on human ratings of the \CleverName{} model, but comparable numbers were also obtained for the KVMemNet model.
The mentioned results follow intuition, as \textit{fluency} of a response is a notion that can be easily defined and identified.
On the other hand \textit{consistency} and \textit{engagingness} are ambiguous, and (possibly) partially overlapping, concepts.

To associate quantitative metrics from Table \ref{tab:hyperp} with human ratings, we computed correlation between perplexity values from the sketch decoder of the \CleverName{} model with human scores across different dimensions.
The study showed no correlation for \textit{fluency} ($\rho = -0.015$), and weak correlations for \textit{consistency} ($\rho = -0.190$) and \textit{engagingness} ($\rho = -0.147$).

\paragraph{Model vocabulary analysis}
To assess the diversity of responses generated by the models, we calculated the percentage of unique $n$-grams and full responses present in the model outputs.
Table \ref{tab:ngram-counts} presents these values for KVMemNet and \CleverName{} computed on the full validation set.
The numbers show that the KVMemNet model clearly outperforms our model in terms of generating diverse and unique outputs by a factor of 3-4x.
However, we hypothesize that this additional diversity may lead to lower engagingness scores.

\begin{figure*}[ht!]
    \centering
    \begin{subfigure}{.3\textwidth}
      \centering
      \includegraphics[width=.87\linewidth]{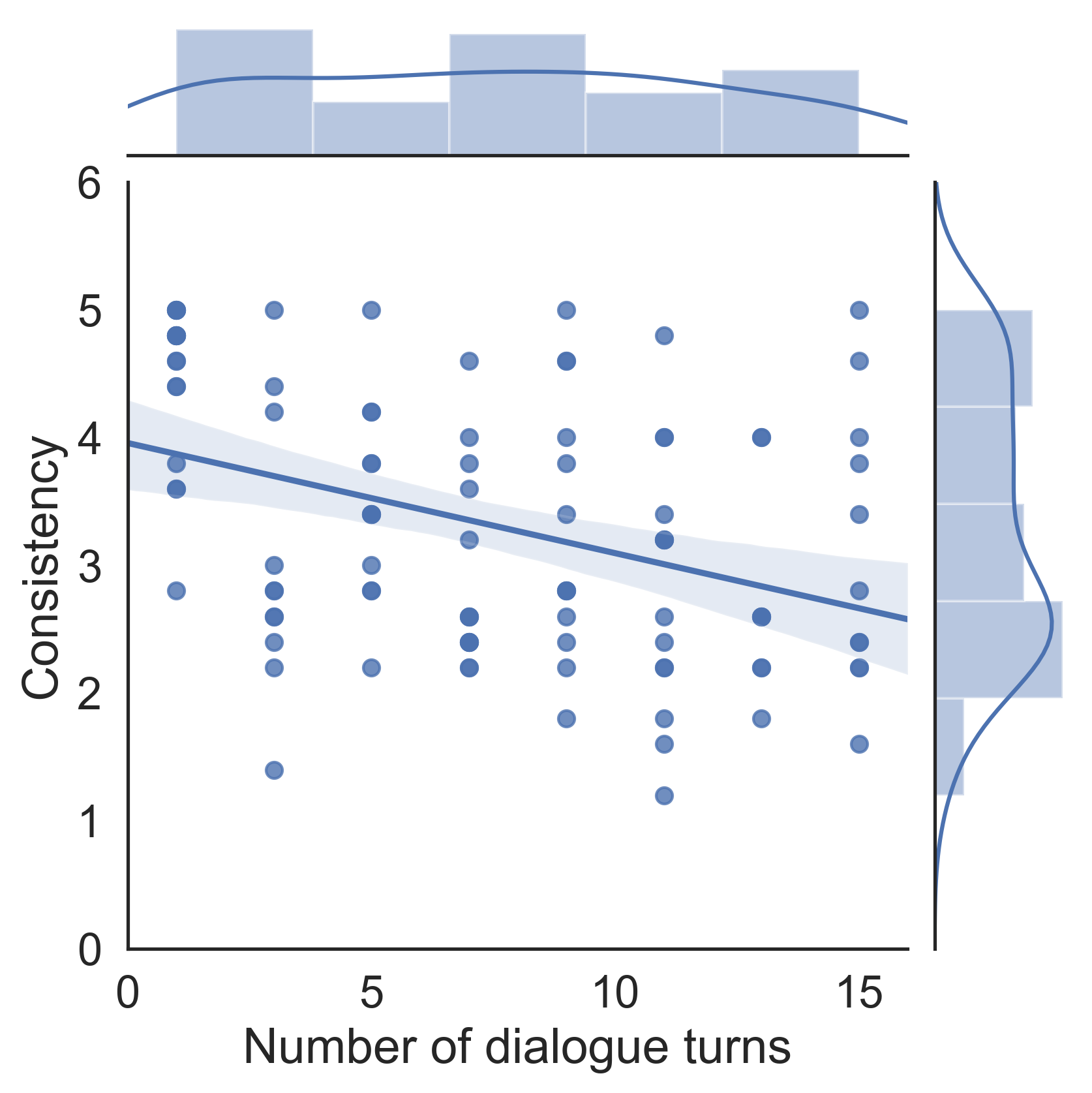}
      \caption{KVMemNet}
      \label{fig:sub1}
    \end{subfigure}%
    \begin{subfigure}{.3\textwidth}
      \centering
      \includegraphics[width=.87\linewidth]{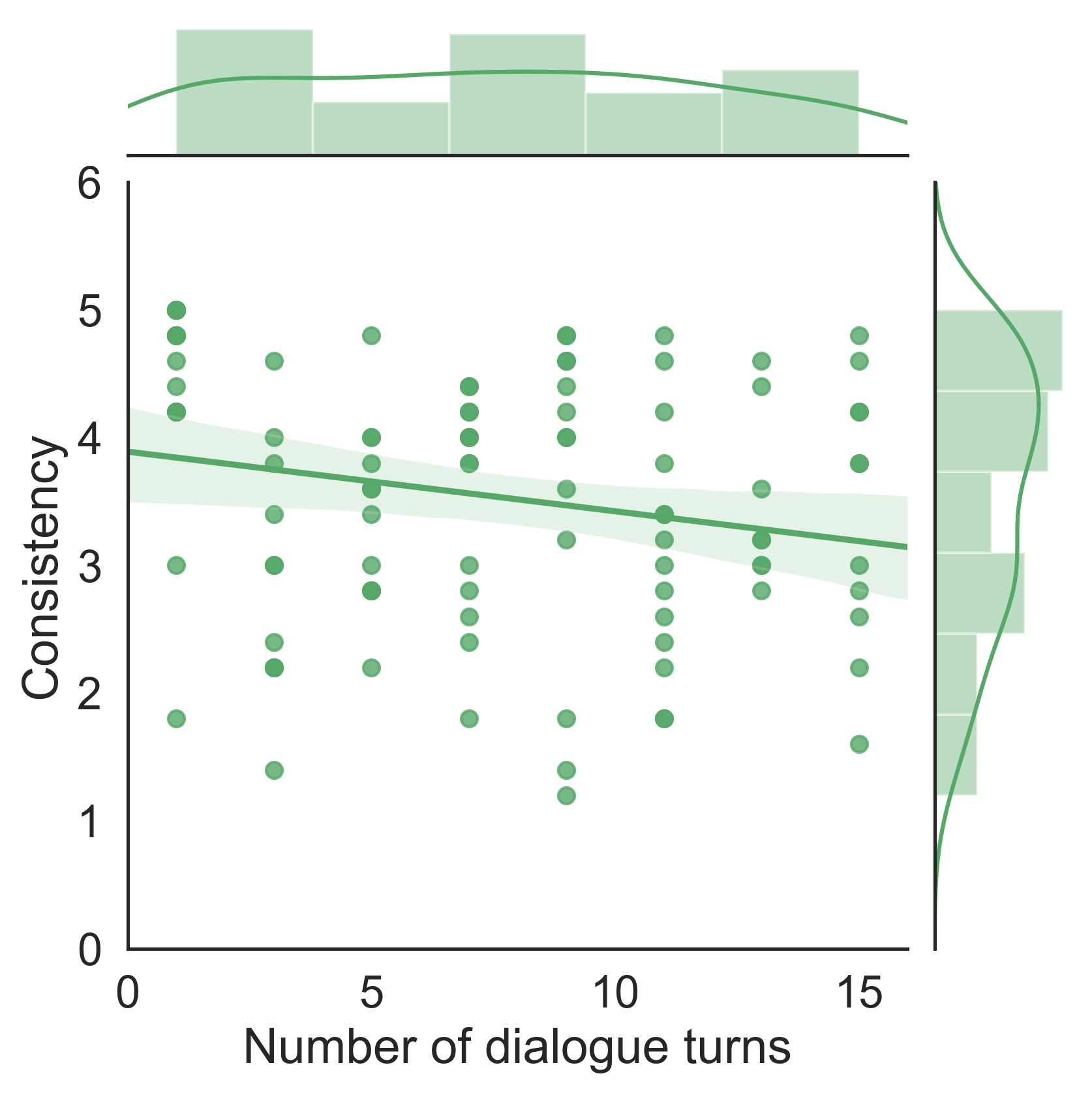}
      \caption{\CleverName{}}
      \label{fig:sub2}
    \end{subfigure}
    \begin{subfigure}{.3\textwidth}
      \centering
       \includegraphics[width=.9\linewidth]{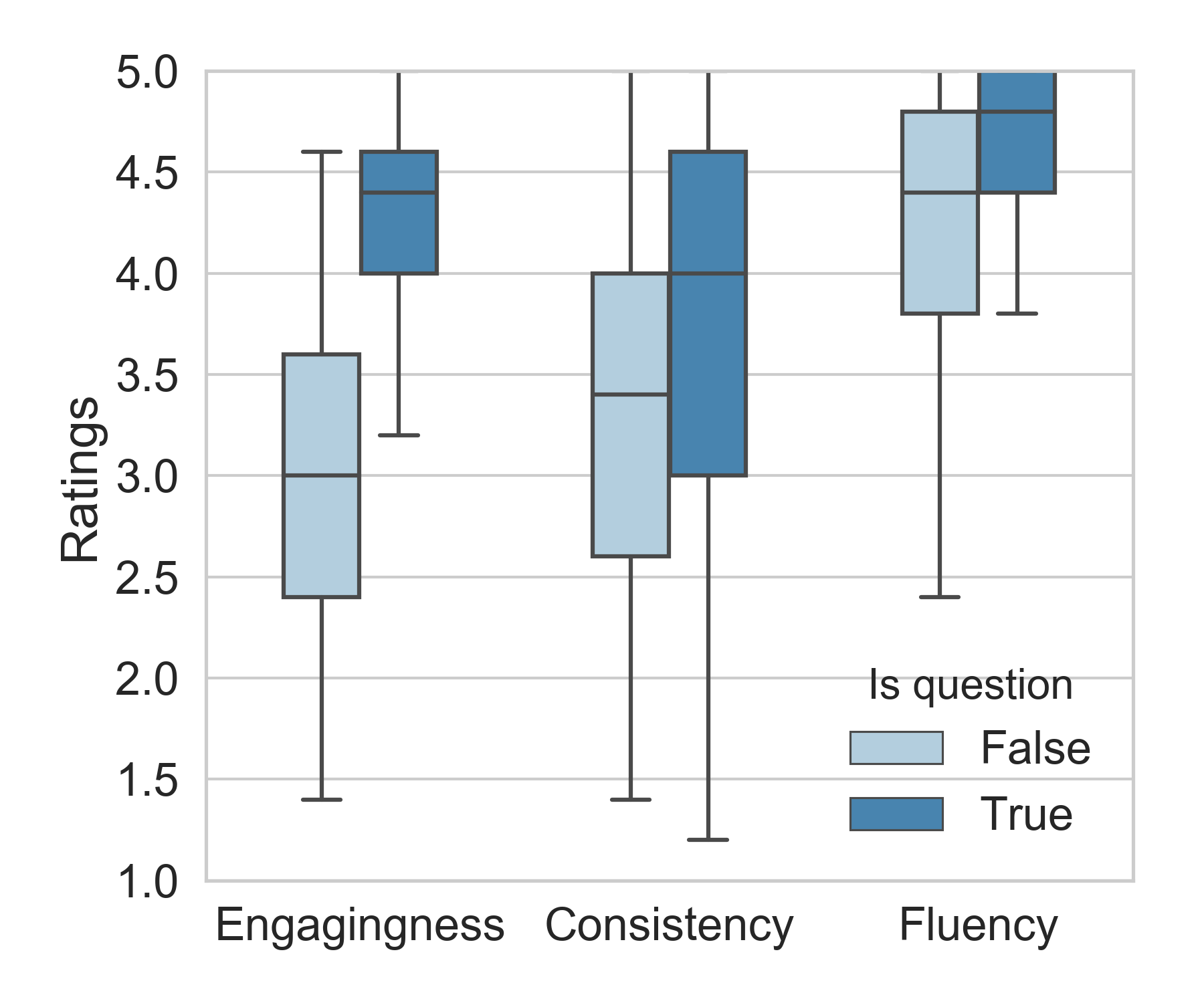}
      \caption{\CleverName{}: Human ratings vs question/no-question responses}
      \label{fig:sub3}
    \end{subfigure}
    \caption{Impact of conversation length on the consistency of outputs generated by the KVMemNet (left) and \CleverName{} (middle). As conversation length increases (more dialogue turns) both models become less consistent, but KVMemNet degrades faster than \CleverName{}.
    Right: impact of response containing a question on human ratings. Responses including questions tend to receive higher human ratings.
    }
    \label{fig:length-influence}
\end{figure*}

\paragraph{Consistency over time}
In order to evaluate the models capacity to stay consistent with its previous statements, and thus implicitly its ability to utilize information present in the chat history, we compared how the consistency rating changed as the number of lines of the conversation increased. 
Figure \ref{fig:length-influence} visualizes this metric both for our model and KVMemNet.
In the case of both models, the consistency decreases as the chat history get longer, indicating that models have problems keeping track of their previous statements. 
When analyzing the linear trend we noticed that the decrease in performance is slower for the \CleverName{} model. 
We hypothesize that this effect can be partially caused by the high diversity of sequences generated by the KVMemNet, which in turn affects the models ability to generate consistent conversation.

\paragraph{Effect of question responses}
We hypothesize that for a conversation to be engaging, responses in chit-chat dialogue should be a mix of statements, where the model shares its persona information, and questions, where the model inquires about certain traits and information of the other agent.
To confirm this intuition, we evaluated the effect that presence of a question in the response has on the ratings coming from the judges.
The results are presented in Figure \ref{fig:sub3}.
The study showed that there is a strong correlation between the model asking a question and the users rating the response as more engaging. 
Asking questions has a small, but positive influence on \textit{engagingness} and \textit{fluency}.

To further analyze this aspect, we measured the frequency of questions in the set of 100 responses coming from the \CleverName{} and KVMemNet models.
We found that our model produced 49 question responses out of which 25 had both a statement and a question. 
In the same setting the KVMemNet produced 15 questions out of which only 1 contained a statement and a question.
This insight could explain the gains on the engagingness ratings found by our human study.

\subsection{Multi-turn User Study}
To evaluate both models in the more challenging multi-turn setting, we collected $30$ conversations that lasted 20 turns, between each model and human users. 
Users were asked to score their conversations with the models on a scale from 1 (lowest) to 5 (highest) across the same dimensions as in the single-turn experiments.
Table \ref{tab:multiturn-human-study} shows the human ratings for both \CleverName{} and KVMemNet. 
Both were judged as less fluent (scores $\approx 3$) than in the single-turn case (scores $\geq 4$).
This is likely due to the models having to respond to a range of conversation histories unseen during training. 

Notably, \CleverName{} outperformed KVMemNet on \textbf{consistency}, by a significantly larger margin \textbf{(3.72 vs 2.15)} than in the single-turn setting.
This suggests that \CleverName{} benefits from conditioning response generation on its persona-memory and so adheres more closely to responses that are compatible with its persona.   

Further, \CleverName{} is more engaging. This suggests that in the multi-turn setting, there also is a positive correlation between engagingness and consistency as in the single-turn case (see Appendix): consistent models can be more engaging as well.

Table \ref{tab:multiturn-kvmemnet} shows an example of KVMemNet's inconsistency. While every model utterance is fluent individually, KVMemNet noticeably contradicts itself in the context of previous utterances and frequently ignores the human responses (e.g "i do not have any myself" after "my little girl"). We believe the lack of structure inherent in models built on vanilla Seq2Seq make KVMemNet prone to this mistake. Table \ref{tab:multiturn-sketchfillar} shows \CleverName{} conducts a more engaging conversation, with pertinent responses and questions. However, this structure can restrict \CleverName{}, as sketches may be filled with incorrect persona traits (e.g "i love papaya food ."). See the Appendix for more  examples.
\section{Discussion and Future Work}

%
%
%
In our study we have identified several paths for future work.
First, our results show that perplexity does not strongly correlate with human judgment of the quality of responses.
Developing an automated metric that correlates well with human judgment is crucial as human evaluation is expensive, time consuming, and prone to inconsistencies.
Secondly, despite outperforming other models in the multi-turn dialogue setting on consistency and engagement, our model has not reached human-like fluency. In order to demonstrate higher-level complex traits such as empathy, models must first master these lower-level abilities. 
Finally, correct usage of rare words and proper nouns leads to higher human scores. Existing models are unable to deal with out-of-vocabulary tokens and rare words gracefully, and incorporation of commonsense via external knowledge bases or other methods will be useful.  



\paragraph{Ethical Implications}
During experiments, we identified a number of ethical implications for future work. 
The \texttt{Persona-Chat} dataset was noted by some raters to contain potentially inappropriate statements (e.g., "my wife spends all my money") and is based in US culture (e.g., food, music, cars, names).
It also lacked content to fail gracefully when it didn't have an appropriate response (e.g., "I'm sorry I don't understand," "I don't know"). 
As such, learned model responses were occasionally insensitive and confusing to human users.

\bibliography{emnlp-ijcnlp-2019}
\bibliographystyle{acl_natbib}

\end{document}